# Fractional spectral graph wavelets and their applications

Jiasong Wu[1, 2, 3, 4], Fuzhi Wu[1], Qihan Yang[1], Youyong Kong[1, 2, 3, 4], Xilin Liu[1, 3, 5], Yan Zhang[1],

Lotfi Senhadji[1, 2, 3, 4], Huazhong Shu[1, 2, 3, 4]

[1]*LIST, the Key Laboratory of Computer Network and Information Integration, Southeast University, Ministry of Education, Nanjing 210096, China*

[2]*International Joint Research Laboratory of Information Display and Visualization, Southeast University, Ministry of Education, Nanjing 210096, China*

[3]*INSERM, U 1099, Université de Rennes 1, LTSI, Rennes 35000, France*

[4]*Centre de Recherche en Information Biomédicale Sino-français, Nanjing 210096, China*

[5]*College of Data Science, Taiyuan University of Technology, Taiyuan 030024, China*

Information about the corresponding author:
Jiasong Wu, Ph.D
LIST, Key Laboratory of Computer Network and Information Integration
School of Computer Science and Engineering
Southeast University, 210096, Nanjing, China
Tel: 00-86-25-83 79 42 49
Fax: 00-86-25-83 79 26 98
Email: jswu@seu.edu.cn

**Abstract**:

One of the key challenges in the area of signal processing on graphs is to design transforms and dictionaries methods to identify and exploit structure in signals on weighted graphs. In this paper, we first generalize graph Fourier transform (GFT) to graph fractional Fourier transform (GFRFT), which is then used to define a novel transform named spectral graph fractional wavelet transform (SGFRWT), which is a generalized and extended version of spectral graph wavelet transform (SGWT). A fast algorithm for SGFRWT is also derived and implemented based on Fourier series approximation. The potential applications of SGFRWT are also presented.

# 1. Introduction

In traditional signal processing, the most commonly used tools are transforms, among which Fourier transform and wavelet transform play key roles [1]. Fourier transform and wavelet transform were generalized and extended in many contexts in recent years.

On the one hand, Fourier transform and wavelet transform were extended to fractional domain, obtaining fractional Fourier transform (FRFT) [2-22] and fractional wavelet transform (FRWT) [23-27]. The FRFT can be interpreted as a rotation in the time-frequency plane, i.e. the unified time-frequency transform. With the order from 0 increasing to 1, the FRFT can show the characteristics of the signal changing from the time domain to the frequency domain [7]. There are roughly three main research directions for investigating FRFT. Firstly, the application of FRFT to deal with many signal processing problems [8], for example, filtering, compression, image encryption, digital watermarking, pattern recognition, edge detection, antennas, radar and sonar, and communication. Secondly, the discretization algorithms of the FRFT [9-16]. Lastly, the extension of the fractional idea of FRFT to other transforms, for example, fractional cosine, sine and Hartley transforms [17-20], fractional Krawtchouk transform [21], short-time FRFT [22]. By cascading of the FRFT and the ordinary wavelet transform, Mendlovic et al. [23] first proposed FRWT. Recently, by introducing a new structure of the fractional convolution [24] associated with the FRFT, Shi et al. [25] proposed a simplified definition of the FRWT which analyzes the signal in time-frequency-FRFD domain, however, the physical meaning of this kind of FRWT requires deeper interpretation [26, 27]. The definition of FRWT in [25] was further improved by Prasad et al. [26], who analyzed the signal only in time-FRFD domain. More recently, Dai et al. [27] presented a new FRWT which displays the time and FRFD-frequency information jointly in the time-FRFD-frequency plane. In general, compared to traditional transforms, fractional transforms may lead to (1) better signal filtering results due to the rotation of the time-frequency plane [8]; (2) better image compression ratios [28], image recognition results [29] and image segmentation results [30] due to the flexible choices of the fractional orders; (3) better image encryption [31] and image watermarking [32] performances since the fractional order can be used as an additional secret key.

On the other hand, Fourier transform and wavelet transform were extended to graph domain,

obtaining graph Fourier transform (GFT) [33-40] and graph wavelet transform (GWT) [41-44] to handle signal defined on the vertices of weighted graphs. Two basic approaches to signal processing on graphs have been considered: The first is rooted in the spectral graph theory [45] and builds upon the graph Laplacian matrix [33]. Since the standard graph Laplacian matrix is restricted to be symmetric and positive semi-definite, this approach is applicable only to undirected graphs with real and nonnegative edge weights. The second approach, discrete signal processing on graphs [37, 38], is rooted in the algebraic signal processing theory [46, 47] and builds on the graph shift operator. This latter works as the elementary operator that generates all linear shift-invariant filters for signals with a given structure. In particular, the graph Fourier transform in this framework expands a graph signal into a basis of eigenvectors of the adjacency matrix, and the corresponding spectrum is then given by the associated eigenvalues. Besides graph-based transforms [33-44], recent research works on graph also include, among others, sampling and interpolation on graphs [48, 49], graph signal recovery [50-53], semi-supervised classification on graphs [54, 55], graph dictionary learning [56, 57], graph convolutional neural networks [58-65]. Please refer to [66] for more references of graph signal processing.

To the authors' knowledge, there are only few research work done for discrete signal processing on fractional graph domain, which is a combination of fractional transform domain and graph transform domain. Wang et al. [67] proposed a definition of fractional Fourier transform on graphs. In this paper, we propose a new spectral graph fractional wavelet transform (SGFRWT), which is an extended version of spectral graph wavelet transform (SGWT) [41].

The remainder of the paper is organized as follows: Section 2 recalls the main foundations of fractional wavelet transform, spectral graph theory and spectral graph wavelets. The spectral graph transforms fractional Fourier transform (SGFRFT) and the spectral graph fractional wavelet transform (SGFRWT) are defined in Section 3. Section 4 is dedicated to the Fourier series approximation based fast algorithm for forward and inverse SGFRWT. Several applications of SGFRWT for different problems are shown in Section 5. Section 6 concludes the paper.

## 2. Preliminary

In the following, we first recall the forward and inverse fractional Fourier transform (FRFT) [2-5], and then the forward and inverse fractional wavelet transform (FRWT) [23, 25-27] and also show how scaling operator may be expressed in the FRFT domain.

*2.1 Fractional Fourier transform (FRFT)*

The $\theta$-order forward FRFT [2-5] is the decomposition of a function $f$ according to the fractional Fourier kernel, i.e.,

$$\hat{f}_\theta(u) = \langle f, K_\theta^* \rangle = \int_{-\infty}^{+\infty} f(x) K_\theta(x,u) dx, \qquad (1)$$

where

$$K_\theta(x,u) = \begin{cases} A_\theta \exp\left[(j/2)(x^2+u^2)\cot\theta - jxu\csc\theta\right], & \theta \neq n\pi \\ \delta(x-u), & \theta = 2n\pi \\ \delta(x+u), & \theta = (2n+1)\pi \end{cases}, \qquad (2)$$

and

$$A_\theta = \sqrt{\frac{1-j\cot\theta}{2\pi}}, \qquad (3)$$

where $\theta$ indicates the rotation angle of the transform and $\delta(x)$ represents the Dirac distribution.

The inverse FRFT is given by [2-5]

$$f(x) = \langle \hat{f}_\theta, K_\theta \rangle = \int_{-\infty}^{+\infty} \hat{f}_\theta(u) K_\theta^*(x,u) du. \qquad (4)$$

The kernel in Eq. (2) can be obtained by its spectral expansion [5],

$$K_\theta(x,u) = \sum_{k=0}^{\infty} \exp(-j\theta k) \xi_k(u) \xi_k(x). \qquad (5)$$

where $\xi_k(x)$ is the $k$th-order normalized Hermite function [10, 11], which is the eigenfunction of the Fourier transform. From (5), we can see that FRFT is a generalized version of the Fourier transform, since they share the same eigenfunction $\xi_k(x)$ and the eigenvalues of the FRFT are the $\theta$-th root of the eigenvalues of the Fourier transform [2]. When the rotation angle $\theta=\pi/2$, the FRFT degrades to Fourier transform.

The discrete form of Eq. (5) can be expressed as follows [10, 11]

$$\mathbf{F}_N^\theta = \mathbf{V}_N \mathbf{D}_N^\theta \mathbf{V}_N^T = \mathbf{V}_N \, diag\left(\left[1, e^{-j\theta}, \ldots, e^{-j(N-1)\theta}\right]\right) \mathbf{V}_N^T, \qquad (6)$$

where *diag*(.) denotes a diagonal matrix formed from its vector argument. $\mathbf{F}_N^\theta$ denotes $\theta$-order inverse FRFT matrix of size $N\times N$. $\mathbf{V}_N = [\mathbf{v}_0 \ \mathbf{v}_1 \ \cdots \ \mathbf{v}_{N-1}]$, $\mathbf{v}_k$ is the $k$th order DFT Hermite eigenvector. Note that $\mathbf{F}_N^\theta$ becomes identity matrix when $\theta=0$ and Fourier matrix when $\theta=\pi/2$.

### 2.2 Fractional wavelet transform (FRWT)

There are several definitions for fractional wavelet transform (FRWT) [23, 25-27]. In this part, we choose the definition in [27] since it displays the signal in the joint time-FRFD-frequency plane and also has a better physical meaning.

The $\theta$-order forward FRWT is defined as [27]

$$W_f^\theta(s,a) = \langle f, \psi_{\theta,s,a} \rangle = \int_{-\infty}^{\infty} f(x)\psi_{\theta,s,a}^*(x)dx, \tag{7}$$

where

$$\psi_{\theta,s,a}(x) = e^{-\frac{j}{2}\left(x^2 - a^2 - \left(\frac{x-a}{s}\right)^2\right)\cot\theta} \psi_{s,a}(x), \tag{8}$$

with $\psi_{s,a}(x) = \frac{1}{s}\psi\left(\frac{x-a}{s}\right)$.

The unitarity property of the FRFT applied to (7) leads to an equivalent form of FRWT:

$$W_f^\theta(s,a) = (T_\theta^s f)(a) = \sqrt{\frac{2\pi}{1-j\cot\theta}} \int_{-\infty}^{\infty} e^{\frac{j}{2}s^2u^2\cot\theta} \hat{f}_\theta(u)\hat{\psi}_\theta^*(su) K_\theta^*(u,a)du, \tag{9}$$

where $\hat{f}_\theta$ stands for the $\theta$-order FRFT of $f$. Here also, as for the classical CWT [1], the scaling is operated in the transformed domain "$u$" and not directly in the time domain.

The inverse of the FRWT is given by [27]

$$f(x) = \frac{1}{2\pi\sin\theta C_{\theta,\psi}} \int_{-\infty}^{\infty}\int_{-\infty}^{\infty} W_f^\theta(s,a)\psi_{\theta,s,a}(x)\frac{dsda}{s}, \tag{10}$$

when the following admissibility condition is satisfied [27]

$$\int_0^\infty \frac{|\hat{\psi}_\theta(s)|^2}{s}ds = C_{\theta,\psi} < \infty, \tag{11}$$

### 2.2 Spectral graph theory and spectral graph wavelets

In this section, we briefly review the spectral graph theory and also the spectral graph wavelets [41].

*2.2.1 Spectral graph theory*

We consider an undirected, connected, weighted graphs $\mathcal{G} = \{\mathcal{V}, \mathcal{E}, \mathbf{W}\}$, composed of a finite set of vertices $\mathcal{V}$ (with $Card(\mathcal{V}) = N$), a set of edges $\mathcal{E}$, and an adjacency symmetric and positive-valued matrix $\mathbf{W}$. A real-valued signal $f$ defined on the vertices $\mathcal{V}$ of the graph $\mathcal{G}$ is a $N \times 1$ vector where each entry is the value $f(n)$ assigned to the vertex $n$. The (non-normalized) graph Laplacian operator is the matrix defined by $\mathbf{L} = \mathbf{D} - \mathbf{W} \in \mathbb{R}^{N \times N}$, where $\mathbf{D}$ is the diagonal degree matrix defined by $d_{m,m} = \sum_n w_{m,n}$. $d_{m,m}$ is the degree of vertex $m$.

As the matrix $\mathbf{L}$ is real and symmetric then it is diagonalizable and its eigenvectors $\{\chi_0, \chi_1, \ldots, \chi_{N-1}\}$, sorted according to ascending order of the corresponding eigenvalues $0 = \lambda_0 < \lambda_1 \leq \lambda_2 \leq \cdots \leq \lambda_{N-1}$ form an orthonormal basis. Therefore, with the unitary column matrix:

$$\boldsymbol{\chi} = [\chi_0, \chi_1, \ldots, \chi_{N-1}], \tag{12}$$

and the diagonal matrix $\boldsymbol{\Lambda} = diag([\lambda_0, \lambda_1, \ldots, \lambda_{N-1}])$ we have:

$$\mathbf{L} = \boldsymbol{\chi} \boldsymbol{\Lambda} \boldsymbol{\chi}^H, \tag{13}$$

where the superscript $H$ denotes the Hermitian transpose operation. The graph Fourier transform (GFT) of the signal $f$ is then the sequence provided by the scalar products [33]:

$$\hat{f}(\ell) = \langle f, \chi_\ell \rangle = \sum_{n=1}^{N} f(n) \chi_\ell^*(n), \quad \ell = 0, 1, \ldots, N-1, \tag{14}$$

and $f$ can be recovered (inverse GFT) by means of the reconstruction formula:

$$f(n) = \langle \hat{f}, \chi_\ell^* \rangle = \sum_{\ell=0}^{N-1} \hat{f}(\ell) \chi_\ell(n), \quad n = 1, \ldots, N. \tag{15}$$

Using vectors and matrices notations, Eqs. (14) and (15) become:

$$\hat{f}_\theta = \begin{bmatrix} \hat{f}_\theta(0) & \hat{f}_\theta(1) & \cdots & \hat{f}_\theta(N-1) \end{bmatrix}^T = \boldsymbol{\chi}^H f, \tag{16}$$

$$f = \begin{bmatrix} f(1) & f(2) & \cdots & f(N) \end{bmatrix}^T = \boldsymbol{\chi} \hat{f}. \tag{17}$$

The graph Fourier transform obeys the Parseval's theorem, that is, for any signals $f$ and $h$ defined on the graph $\mathcal{G}$ we have:

$$\langle f, h \rangle = \langle \hat{f}, \hat{h} \rangle. \tag{18}$$

*2.2.2 Spectral graph wavelet transform (SGWT)*

The spectral graph wavelet transform (SGWT) of the signal $f$ with the kernel $g$ is defined by [41]:

$$W_f(s,n) = (T_g^s f)(n) = \sum_{\ell=0}^{N-1} g(s\lambda_\ell) \hat{f}(\ell) \chi_\ell(n), \quad n = 1,...,N, \tag{19}$$

where

$$T_g^s = g(t\mathbf{L}), \tag{20}$$

and the kernel $g$ is continuous positive-valued function defined on $\mathbb{R}^+$ satisfying:

$$g(0) = \lim_{x \to +\infty} g(x) = 0 \, ; \quad \int_0^{+\infty} \frac{g(x)}{x^2} dx = C_g \in \mathbb{R}^+. \tag{21}$$

Using Eq. (14), the SGWT becomes

$$W_f(s,n) = (T_g^s f)(n) = \sum_{m=1}^{N} f(m) \psi_{s,n}^*(m) = \langle f, \psi_{s,n} \rangle, \quad n = 1,...,N, \tag{22}$$

with

$$\psi_{s,n}(m) = \sum_{\ell=0}^{N-1} g(s\lambda_\ell) \chi_\ell(m) \chi_\ell^*(n), \quad m = 1,...,N. \tag{23}$$

The signal $f$ can be recovered up to its mean value using the inverse formula [41]:

$$f(m) - \langle f, \chi_0 \rangle \chi_0(m) = \frac{1}{C_g} \sum_{n=1}^{N} \int_0^\infty W_f(s,n) \psi_{s,n}(m) \frac{ds}{s}, m = 1,...,N. \tag{24}$$

## 3. Spectral graph fractional Transforms

We first propose a novel spectral graph fractional Fourier transform (GFRFT) in Section 3.1, and then give the definition of spectral graph fractional wavelet transform (SGFRWT) in Section 3.2. Finally, we give some properties of SGFRWT in Section 3.3.

### *3.1 Spectral Graph fractional Fourier Transform*

The forward graph fractional Fourier transform (GFRFT) of any signal $f$ defined on the vertices $\mathcal{V}$ of the graph $\mathcal{G}$ is defined by:

$$\hat{f}_\theta(\ell) := \langle f, \chi_\ell \rangle = \sum_{n=1}^{N} f(n) \gamma_\ell^*(n), \quad \ell = 0,1,...,N-1, \tag{25}$$

or by the following matrix form:

$$\hat{f}_\theta = \begin{bmatrix} \hat{f}_\theta(0) & \hat{f}_\theta(1) & \cdots & \hat{f}_\theta(N-1) \end{bmatrix}^T = \boldsymbol{\gamma}^H f, \tag{26}$$

where $0 < \theta \leq 1$ and $\boldsymbol{\gamma}$ is the power matrix given by:

$$\boldsymbol{\gamma} = [\gamma_0, \gamma_1, \ldots, \gamma_{N-1}] = \boldsymbol{\chi}^\theta, \tag{27}$$

with $\boldsymbol{\chi}$ being the unitary (i.e. inverse graph Fourier transform matrix) shown in (12). Note that as $\boldsymbol{\chi}$ is unitary then $\boldsymbol{\gamma} = \boldsymbol{\chi}^\theta$ is also unitary.

For $\theta = 1$, the proposed GFRFT in (25) is no more than the GFT in (18) defined by Shuman et al. [33]. For other values of $\theta$, we can compute $\boldsymbol{\gamma} = \boldsymbol{\chi}^\theta$ by using schur–padé algorithm [68, 69].

The *inverse GFRFT* is given by:

$$f(n) = \langle \hat{f}_\theta, \gamma_\ell^* \rangle = \sum_{\ell=0}^{N-1} \hat{f}_\theta(\ell) \gamma_\ell(n), \quad n = 1, \ldots, N, \tag{28}$$

or by the following matrix form

$$f = \begin{bmatrix} f(1) & f(2) & \cdots & f(N) \end{bmatrix}^T = \boldsymbol{\gamma} \hat{f}_\theta. \tag{29}$$

The graph fractional Fourier transform obeys the Parseval's theorem, that is, for any signals $f$ and $h$ defined on the graph $\mathcal{G}$ we have:

$$\langle f, h \rangle = \langle \hat{f}_\theta, \hat{h}_\theta \rangle. \tag{30}$$

*3.2 Spectral graph fractional wavelet transform (SGFRWT)*

Similar to (13), we define the graph fractional Laplacian operator $\mathbf{L}_\theta$ as follows

$$\mathbf{L}_\theta = \boldsymbol{\gamma} \mathbf{R} \boldsymbol{\gamma}^H, \tag{31}$$

where $\boldsymbol{\gamma}$ is defined in (27), and

$$\mathbf{R} = diag([r_0, r_1, \ldots, r_{N-1}]) = \boldsymbol{\Lambda}^\theta, \tag{32}$$

that is,

$$r_\ell = \lambda_\ell^\theta, \quad \ell = 0, 1, \ldots, N-1. \tag{33}$$

Considering $\lambda_\ell \in [0, \lambda_{N-1}]$, and from (33), we can easily get $r_\ell \in [0, r_{N-1}]$.

Similar to (22), we can define the spectral graph fractional wavelet transform (SGFRWT) operator $T_{g_\theta}^s$ as follows

$$W_f(\theta,s,n) = \left(T_{g_\theta}^s f\right)(n) = \sum_{m=1}^{N} f(m)\psi_{\theta,s,n}^*(m) = \langle f, \psi_{\theta,s,n}\rangle, \quad n=1,...,N, \quad (34)$$

where

$$T_{g_\theta}^s = g(s\mathbf{L}_\theta), \quad (35)$$

$$\psi_{\theta,s,n}(m) = \sum_{\ell=0}^{N-1} g(s\lambda_\ell^\theta)\gamma_\ell(m)\gamma_\ell^*(n), \quad m=1,...,N. \quad (36)$$

From (34), we can see that the proposed SGFRWT is a generalization of SGWT in (22) defined by Hammond et al. [41], that is, when $\theta=1$, the proposed SGFRWT degrades to the SGWT.

The scaling functions are then obtained by $\phi_\theta(n) = T_{h_\theta}\delta_n = h(\mathbf{L}_\theta)\delta_n$, and the coefficients by $S_{\theta,f}(n) = \langle \phi_{\theta,n}, f\rangle$. Stable recovery of the original signal **f** from the SGFRWT coefficients will be assured if the quantity $G(r) = h^2(r) + \sum_{j=1}^{J} g^2(t_j r)$ is bounded away from zero on the fractional Laplacian $\mathbf{L}_\theta$. Moreover, under the same conditions, lemmas 5.1, 5.2, 5.3, 5.4 and theorems 5.5 and 5.6 in [41] are verified by the SGFRWT.

## 4. Fourier series approximation and fast SGFRWT

If we compute the SGFRWT by using the Eq. (34) directly, we have to compute the entire set of eigenvectors and eigenvalues of $\mathbf{L}_\theta$ shown in (31). Both the computational complexity and the memory consumption become unacceptable when the data is larger than hundreds of thousands or millions of dimensions. Therefore, in this section, we propose a fast algorithm for computing the SGFRWT based on approximating the scaled generating kernels $g_\theta$ by low order Fourier series. It should be noted that the proposed fast SGFRWT algorithm is an extension of the fast SGWT algorithm from real domain to complex domain. The proposed fast SGFRWT algorithm is different from the fast graph Fourier transform proposed by Le Magoarou *et al.* [36], who use the modified Jacobi eigenvalue algorithm for approximating the graph Laplacian matrix. Note that the proposed SGFRWT algorithm can approximate an arbitrary matrix while the algorithm in [36] is only practicable for approximating symmetric matrix.

The following Lemma shows that the polynomial approximation of $g(tx)$ may be taken over a finite range containing the spectrum of $\mathbf{L}_\theta$.

**Lemma 4.1.** Let $r_{\max} \geq r_{N-1}$ be any upper bound on the fractional spectrum of $\mathbf{L}_\theta$. For fixed $t > 0$, let $p(x)$ be a polynomial approximant of $g(tx)$ with $L_\infty$ error $B = \sup_{x \in [0, r_{\max}]} |g(tx) - p(x)|$. Then the approximate SGFRWT coefficients $\tilde{W}_f(\theta, t, n) = (p(\mathbf{L}_\theta) f)_n$ satisfy $|W_f(\theta, t, n) - \tilde{W}_f(\theta, t, n)| \leq B \|f\|$.

The proof of Lemma 4.1 is similar to Lemma 6.1 in [41].

In [41], Hammond et al. approximated $g(s\mathbf{L})$ shown in (20) by truncated Chebyshev expansions. The reason is that $g(s\mathbf{L})$ is real as the original $\mathbf{L}$ shown in (13) is a real matrix. However, the Chebyshev polynomial approximation method is not suitable for our problem of approximating $g(s\mathbf{L}_\theta)$ shown in (35), because $g(s\mathbf{L}_\theta)$ may be complex as $\mathbf{L}_\theta$ shown in (31) may be a complex matrix. Therefore, in this part, we will use the truncated Fourier series expansion instead of Chebyshev expansions to approximate $g(s\mathbf{L}_\theta)$ shown in (35).

The Fourier series of a general function $f(x)$ is given by [70]

$$f(y) = \sum_{k=-\infty}^{\infty} c_k \exp(iky), \tag{37}$$

where

$$c_k = \frac{1}{2\pi} \int_0^{2\pi} f(y) \exp(-iky) dy = \begin{cases} a_0, & k = 0 \\ (a_k - ib_k)/2, & k > 0 \\ (a_k + ib_k)/2, & k < 0 \end{cases}, \tag{38}$$

$$a_0 = \frac{1}{2\pi} \int_0^{2\pi} f(y) dy, \quad a_k = \frac{1}{\pi} \int_0^{2\pi} f(y) \cos(ky) dy, \quad b_k = \frac{1}{\pi} \int_0^{2\pi} f(y) \sin(ky) dy. \tag{39}$$

We now assume a fixed set of wavelet scales $t_n$. For each $n$, approximating $g(t_n x)$ for $x \in [0, r_{\max}]$ can be done by shifting the domain of $\exp(-iky)$, $y \in [0, 2\pi]$, using the transformation $x = y r_{\max}/(2\pi)$. Then, $g(t_n x)$ can be written by

$$g(t_n x) = \sum_{k=-\infty}^{\infty} c_{n,k} F_k(x), \quad x \in [0, r_{\max}], \tag{40}$$

with

$$c_{n,k} = \frac{1}{2\pi r_{\max}} \int_{-\pi}^{\pi} g(t_n x) F_{-k}(x) dx, \tag{41}$$

$$F_k(x) = \exp(i2\pi kx / r_{\max}). \tag{42}$$

Note that

$$F_k(x) = (F_{-k}(x))^*. \tag{43}$$

The above equation is due to the fact that

$$\int_0^{r_{\max}} F_k(x) F_{-l}(x) \frac{1}{r_{\max}} dx = \delta_{kl}. \tag{44}$$

For each scale $t_j$, $g(t_j x)$ can be approximated by the truncated Fourier expansion (37) to $2M_j + 1$ terms, that is,

$$p_j(x) = \sum_{k=-M_j}^{M_j} c_{n,k} F_k(x), \quad x \in [0, r_{\max}]. \tag{45}$$

We may choose an appropriate $M_j$ in practice to achieve a balance between accuracy and computational complexity. We can also use exactly the same scheme to approximate the scaling function kernel $h_\theta$ by $p_0$. Then, the approximated SGFRWT wavelet and scaling function coefficients are respectively given by

$$\tilde{W}_f(\theta, t_j, n) = \left( \sum_{k=-M_j}^{M_j} c_{j,k} F_k(\mathbf{L}_\theta) \mathbf{f} \right)_n, \tag{46}$$

$$\tilde{S}_f(\theta, n) = \left( \sum_{k=-M_0}^{M_0} c_{0,k} F_k(\mathbf{L}_\theta) \mathbf{f} \right)_n, \tag{47}$$

where

$$F_k(\mathbf{L}_\theta) = \exp(i2\pi k \mathbf{L}_\theta / r_{\max}). \tag{48}$$

Note that

$$F_k(\mathbf{L}_\theta) = (F_{-k}(\mathbf{L}_\theta))^*. \tag{49}$$

The efficiency of this algorithm depends on the following recursive formula

$$F_k(\mathbf{L}_\theta) \mathbf{f} = F_1(\mathbf{L}_\theta)(F_{k-1}(\mathbf{L}_\theta) \mathbf{f}), \quad k \in [-M_j, M_j]. \tag{50}$$

As the signal $\mathbf{f} \in \mathbb{R}^N$, taking the Eq. (49) into consideration, Eq. (50) can only be computed for $k \in [1, M_j]$, and the other $k \in [-M_j, -1]$ can be obtained by conjugate operation.

The computational complexity of the proposed fast SGFRWT algorithm is analyzed as follows:

1) The computation of $\boldsymbol{\gamma}$ from $\boldsymbol{\chi}$ in (27) by using schur–padé algorithm [68, 69] needs $O(N^3)$ operations.

2) The computation cost of each $F_k(\mathbf{L}_\theta)\mathbf{f}$ in (50) is $O(6|E|)$, where $|E|$ is the number of nonzero edges in the graph, and $6|E|$ is because $F_1(\mathbf{L}_\theta)$ is a complex matrix and a complex multiplication requires 4 real multiplications and 2 real additions (6 operations). Therefore, the computation of all $F_k(\mathbf{L}_\theta)\mathbf{f}, k=1,2,...,M_j$ needs $O(6M_j|E|)$ operations.

3) It requires $O(6N)$ operations at scale $j$ for each $k \leq M_j$ in (46) and (47). Therefore, the computation of (46) and (47) needs $O\left(6N\sum_{j=0}^{J}M_j\right)$ operations.

Therefore, the total computational complexity of the fast SGFRWT algorithm is

$$O\left(N^3 + 6M_j|E| + 6N\sum_{j=0}^{J}M_j\right).$$

Implementation of (50) requires memory of size $2N$. The total memory requirement for the proposed fast SGFRWT is $2N(J+1)+2N$.

*4.1. Fast computation of adjoint*

Given a fixed set of wavelet scales $\{t_j\}_{j=1}^{J}$, and including the scaling functions $\phi_n$, the SGFRWT $\mathbf{W}_\theta f = \left((T_{h_\theta}\mathbf{f})^T, (T_{g_\theta}^{t_1}\mathbf{f})^T, ..., (T_{g_\theta}^{t_J}\mathbf{f})^T\right)^T$ can be seen as a linear map $\mathbf{W}_\theta : \mathbb{R}^N \to \mathbb{C}^{N(J+1)}$, where $\mathbb{C}$ denotes the complex domain. Let $\tilde{\mathbf{W}}_\theta f = \left(((p_0(\mathbf{L}_\theta))\mathbf{f})^T, (p_1(\mathbf{L}_\theta)\mathbf{f})^T, ..., (p_J(\mathbf{L}_\theta)\mathbf{f})^T\right)^T$ denote the approximated fractional wavelet transform by using the fast SGFRWT algorithm. In this section, we show that both the adjoint $\mathbf{W}_\theta^* : \mathbb{C}^{N(J+1)} \to \mathbb{R}^N$ and the composition $\mathbf{W}_\theta^*\mathbf{W}_\theta : \mathbb{R}^N \to \mathbb{R}^N$ can be computed efficiently by using the Fourier series approximation.

For any $\boldsymbol{\eta} \in \mathbb{C}^{N(J+1)}$, we consider $\boldsymbol{\eta}$ as the concatenation $\boldsymbol{\eta} = \left(\boldsymbol{\eta}_0^T, \boldsymbol{\eta}_1^T, ..., \boldsymbol{\eta}_J^T\right)^T$ with each $\boldsymbol{\eta}_j \in \mathbb{C}^N$ for $0 \leq j \leq J$, we have

$$\langle \boldsymbol{\eta}, \mathbf{W}_\theta \mathbf{f} \rangle_{N(J+1)} = \langle \boldsymbol{\eta}_0, T_{\theta,h} \mathbf{f} \rangle + \sum_{j=1}^{J} \langle \boldsymbol{\eta}_j, T_{\theta,g}^{t_j} \mathbf{f} \rangle_N = \langle T_{\theta,h}^* \boldsymbol{\eta}_0, \mathbf{f} \rangle + \left\langle \sum_{j=1}^{J} \left(T_{\theta,g}^{t_j}\right)^* \boldsymbol{\eta}_j, \mathbf{f} \right\rangle_N = \left\langle T_{\theta,h}^* \boldsymbol{\eta}_0 + \sum_{j=1}^{J} \left(T_{\theta,g}^{t_j}\right)^* \boldsymbol{\eta}_j, \mathbf{f} \right\rangle_N. \quad (51)$$

Considering $\langle \boldsymbol{\eta}, \mathbf{W}_\theta \mathbf{f} \rangle_{N(J+1)} = \langle \mathbf{W}_\theta^* \boldsymbol{\eta}, \mathbf{f} \rangle_N$, we have

$$\mathbf{W}_\theta^* \boldsymbol{\eta} = T_{\theta,h}^* \boldsymbol{\eta}_0 + \sum_{j=1}^{J} \left(T_{\theta,g}^{t_j}\right)^* \boldsymbol{\eta}_j. \quad (52)$$

Similar to (46) and (47), the adjoint operator $\mathbf{W}_\theta^*$ in (52) can be approximated by

$$\tilde{\mathbf{W}}_\theta^* \boldsymbol{\eta} = \sum_{j=0}^{J} p_j^*(\mathbf{L}_\theta) \boldsymbol{\eta}_j. \quad (53)$$

In the following, we derive a fast algorithm for the computation of $\tilde{\mathbf{W}}_\theta^* \tilde{\mathbf{W}}_\theta$, which is an approximation of the composition $\mathbf{W}^* \mathbf{W}$. In general, when computing $\tilde{\mathbf{W}}_\theta^* \tilde{\mathbf{W}}_\theta$, we first apply $\tilde{\mathbf{W}}_\theta$, and then apply $\tilde{\mathbf{W}}_\theta^*$ by the fast SGFRWT, which needs $2J$ Fourier series expansions. However, the computational complexity of $\tilde{\mathbf{W}}_\theta^* \tilde{\mathbf{W}}_\theta$ can further be reduced. Note that

$$\tilde{\mathbf{W}}_\theta^* \tilde{\mathbf{W}}_\theta \mathbf{f} = \sum_{j=0}^{J} p_j^*(\mathbf{L}_\theta)\left(p_j(\mathbf{L}_\theta)\mathbf{f}\right) = \sum_{j=0}^{J} \left(p_j^*(\mathbf{L}_\theta) p_j(\mathbf{L}_\theta)\right) \mathbf{f}. \quad (54)$$

In the following, we derive a fast algorithm for the computation of $\tilde{\mathbf{W}}_\theta^* \tilde{\mathbf{W}}_\theta f$.

**Lemma 4.2.** Set $P(x) = \sum_{j=0}^{J} p_j^*(x) p_j(x)$, which has degree $M^* = 2\max\{M_j\}$. Then, $P(x)$ can be computed by

$$P(x) = \sum_{k=-M^*}^{M^*} d_k F_k(x), \quad (55)$$

where $F_k(x) = \exp\left(i\frac{2\pi k x}{r_{\max}}\right)$, $d_k = \sum_{j=0}^{J} d_{j,k}$, and $d_{j,k} = \begin{cases} \sum_{i=-M_j-i}^{M_j} c_{j,i}^* c_{j,k+i}, & -2M_j \leq k \leq -1 \\ \sum_{i=-M_j}^{M_j-k} c_{j,i}^* c_{j,k+i}, & 0 < k \leq 2M_j \end{cases}$.

**Proof.**

$$p_j^*(x)p_j(x) = \left(\sum_{k=-M_j}^{M_j} c_{j,k} F_k(x)\right)^* \left(\sum_{l=-M_j}^{M_j} c_{j,l} F_l(x)\right) = \sum_{k=-M_j}^{M_j} \sum_{l=-M_j}^{M_j} c_{j,k}^* c_{j,l} F_k^*(x) F_l(x) = \sum_{k=-M_j}^{M_j} \sum_{l=-M_j}^{M_j} c_{j,k}^* c_{j,l} F_{l-k}(x)$$

$$= \sum_{k=-M_j}^{M_j} \sum_{k'+k=-M_j}^{M_j} c_{j,k}^* c_{j,k'+k} F_{k'}(x) = \sum_{i=-M_j}^{M_j} \sum_{k+i=-M_j}^{M_j} c_{j,i}^* c_{j,k+i} F_k(x) = \sum_{i=-M_j}^{M_j} \sum_{k=-M_j-i}^{M_j-i} c_{j,i}^* c_{j,k+i} F_k(x)$$

$$= \sum_{k=-2M_j}^{-1} \sum_{i=-M_j-i}^{M_j} c_{j,i}^* c_{j,k+i} F_k(x) + \sum_{k=0}^{2M_j} \sum_{i=-M_j}^{M_j-k} c_{j,i}^* c_{j,k+i} F_k(x) = \sum_{k=-M^*}^{M^*} d_{j,k} F_k(x)$$

Note that $F_k^*(x)F_l(x) = F_{l-k}(x)$ is used in the above derivation.

Therefore,

$$P(x) = \sum_{j=0}^{J} p_j^*(x)p_j(x) = \sum_{j=0}^{J} \sum_{k=-M^*}^{M^*} d_{j,k} F_k(x) = \sum_{k=-M^*}^{M^*} \sum_{j=0}^{J} d_{j,k} F_k(x) = \sum_{k=-M^*}^{M^*} d_k F_k(x).$$

The proof of Lemma 4.2 is now completed. □

Combining (54) and (55), we have

$$\tilde{\mathbf{W}}_\theta^* \tilde{\mathbf{W}}_\theta \mathbf{f} = P(\mathbf{L}_\theta)\mathbf{f} = \sum_{k=-M^*}^{M^*} d_k F_k(\mathbf{L}_\theta)\mathbf{f}. \tag{56}$$

From (56), we can see that $\tilde{\mathbf{W}}_\theta^* \tilde{\mathbf{W}}_\theta \mathbf{f}$ can be computed by a single Fourier series expansion with twice the length, which reduces the computational cost by a factor $J$ comparing to the direct computation of $\tilde{\mathbf{W}}_\theta^* \tilde{\mathbf{W}}_\theta \mathbf{f}$.

*4.2. Reconstruction by using the fast adjoint algorithm*

In this part, we show the reconstruction of proposed SGFRWT by using the fast adjoint algorithm in Section 4.1.

The SGFRWT is an overcomplete transform since it maps an input signal $\mathbf{f}$ of size $N$ to the $N(J+1)$ factional wavelets coefficients

$$\mathbf{c} = \mathbf{W}_\theta \mathbf{f}. \tag{57}$$

The pseudoinverse of the above equation can be obtained by solving the following square matrix equation

$$\left(\mathbf{W}_\theta^* \mathbf{W}_\theta\right)\mathbf{f} = \mathbf{W}^* \mathbf{c}. \tag{58}$$

Eq. (58) can be solved by the conjugate gradients method [71], whose computation in each step is dominated by applying $\mathbf{w}_\theta^* \mathbf{w}_\theta$ to a single vector. What's more, $\mathbf{w}_\theta^* \mathbf{w}_\theta \mathbf{f}$ can be fast approximated by $\tilde{\mathbf{w}}_\theta^* \tilde{\mathbf{w}}_\theta \mathbf{f}$ shown in (56).

## 5. Application examples

Similar to [41], in this section we will show several application examples of the SGFRWT on different real and synthetic data sets by modifying the GSPBOX toolbox [72]. The first and the second experiments are implemented in Matlab programming language on a PC machine, which sets up Microsoft Windows 7 operating system and has an Intel(R) Core(TM) i7-2600 CPU with speed of 3.40 GHz and 16 GB RAM. The third experiment is implemented using Tensorflow on a PC machine, which sets up Ubuntu 16.04 operating system and has an Intel(R) Core(TM) i7-2600 CPU with speed of 3.40 GHz and 64 GB RAM, and has also two NVIDIA GTX1080-TI GPUs.

5.1 SGFRWT design details

We should design two function kernels in SGFRWT. One is the wavelet general kernel $g(x)$, and the other is the scaling function kernel $h(x)$.

For the wavelet general kernel $g(x)$, we choose the same cubic spline in [41] as follows:

$$g(x) = \begin{cases} x_1^{-\alpha} x^\alpha & \text{for } x < x_1 \\ s(x) & \text{for } x_1 \leq x \leq x_2 \\ x_2^{\beta} x^{-\beta} & \text{for } x > x_2 \end{cases} \quad (59)$$

where $\alpha$ and $\beta$ are integers and they are used to determine the decay rate of spline. $x_1$ and $x_2$ are boundary of transitions region. Note that $g$ is normalized such that $g(x_1)=g(x_2)=1$. The coefficients of the cubic polynomial $s(x)$ are determined by the $s(x_1)=s(x_2)=1$, $s'(x_1)=\alpha/x_1$, and $s'(x_2)=-\beta/x_2$. All of the examples in this paper were produced using $\alpha = \beta = 2$, $x_1 = 1$ and $x_2 = 2$; in this case $s(x)=-5+11x-6x^2+x^3$. For the SGFRWT, we can easily substitute $x$ in (59) by the $\mathbf{L}_\theta$ shown in (31) to get $g(\mathbf{L}_\theta)$.

For the scaling function kernel $h(x)$, we take $h(x) = \rho \exp\left(-\left(\dfrac{x}{0.6\lambda_{\min}}\right)^4\right)$. When $x$ is set to be 0, $\rho$ can be obtained immediately, $\rho = h(0)$.

5.2 SGFRWT application examples

5.2.1 SGFRWT on a synthetic graph data

In the first example, we perform the proposed SGFRWT on "Swiss roll", which is a data set whose points are sampled randomly from a two-dimensional (2D) manifold as follows

$$\bar{x}(s,t) = (t\cos(t)/4\pi, s, t\sin(t)/4\pi), \quad -1 \leq s \leq 1, \; \pi \leq t \leq 4\pi. \tag{60}$$

We take 500 points sampled uniformly on the manifold, and then build a weighted graph, whose edge weight is defined as follows

$$a_{i,j} = \exp(-\|x_i - x_j\|^2 / 2\sigma^2). \tag{61}$$

We use σ = 0.1 for computing the underlying weighted graph, and $J = 4$ scales with $K = 20$ for computing the spectral fractional graph wavelets. Fig. 1 represents the Swiss roll data set, and some SGFRWT coefficients at four different scales localized at the same location. The results of performing SGFRWT with fractional order $\theta$=0.1 on Swiss roll data cloud with $j$=1,2,3,4 wavelet scales are shown in figure (0.1, 0) to (0.1, 4), respectively. From the figure (0.1, 4) to (1.0, 4), we can see that the support of the coarse scale wavelets diffuses along the manifold more locally when the fractional order increase from 0.1 to 1.0.

5.2.2 SGFRWT on a real image

In the second example, we perform the proposed SGFRWT on a standard image "cameraman". We first construct the graph from the 64×64 cameraman image. The vertices of the graph are made up of pixels in the image and each pixel is connected with its vertical and horizontal neighbor pixel. The edge weight is calculated by a Gauss kernel weighting function [33] as follows:

$$w_{i,j} = \begin{cases} \exp\left(-\dfrac{[dist(i,j)]^2}{2\theta^2}\right) & \text{if } dist(i,j) \leq k \\ 0 & \text{otherwise} \end{cases} \tag{62}$$

where $w_{i,j}$ is the weight on the edge connection of the vertex $v_i$ and $v_j$, and $dist\,(i, j)$ represents the distance between pixel $i$ and pixel $j$. We use $J = 5$ scales with $K = 20$ for computing the spectral fractional graph wavelets. Fig. 2 represents the SGFRWT coefficients with fractional order θ ∈ [0.0,1.0] at different wavelet scales $j\in[0,5]$ localized at the same location. From the Figs. (θ, 0) to (θ, 5) for a fixed θ∈[0.0,1.0], we can see that the proposed SGFRWT can analyze the cameraman image from the coarse scale to the fine scale. From the Figs. (0.0, $j$) to (1.0, $j$) for a fixed $j\in[0, 5]$, we can see that the supports of the SGFRWT are more locally and the edges of the cameraman image are more clear when the fractional order increase from 0.1 to 1.0.

As we can see from the two examples, the proposed SGFRWT is a generalization of SGWT [41] since SGFRWT introduces an additional parameter (fractional order $\theta$), which can support a more flexible analysis of graph signals at different fractional domains.

5.2.3 SGFRWT on MNIST database

In the third example, we will see that the proposed SGFRWT can be used as a data augmentation method in the preprocessing of convolutional neural networks. Because, as shown in the second example, the proposed SGFRWT allows deriving many images from the original image by considering several fractional orders. In the following, a series of experiments are performed on the MNIST dataset to verify the effectiveness of the proposed SGFRWT as a method of data augmentation.

The original MNIST database has a training set of 60,000 images and a testing set of 10,000 images, which are then augmented by the following four strategies: (i) For each image of the MNIST training set and testing set, we perform SGWT [41] with $J = 5$ wavelet scales and generate a total of 420,000 images, of which 360,000 images were used for training and 60,000 for testing. (ii) For each image of the MNIST training set and testing set, we apply the proposed SGFRWT with $J = 5$ wavelet scales and fractional order $\theta \in \{0.2, 0.4, 0.6, 0.8, 1.0\}$, generating a total of 2.1 million images. (iii) We apply SGWT [41] on each image of the MNIST training set, and apply the proposed SGFRWT with $J = 5$ wavelet scales on each image of the MNIST testing set. To evaluate the experimental results fairly, we pick up randomly 60,000 images from the expanded testing set so that the total number of images is exactly the same as in case (i). (iv) We apply SGWT [41] on each image of the MNIST training set, and then use traditional data augmentation methods such as flipping, rotating and adding noise to generate a total number of 1.8 million training images, which are the same as case (ii). We then apply SGFRWT with $J = 5$ wavelet scales on each image of the MNIST testing set. The details of the generated four datasets are shown in Table 1. Fig. 3 shows all sample images by processing an image of the digit "zero" by SGFRWT with $J = 5$ wavelet scales and fractional order $\theta \in \{0.2, 0.4, 0.6, 0.8, 1.0\}$.

To evaluate the data augmentation performance of the proposed SGFRWT, we train and test a simple convolutional neural network, which is shown in Fig.4, using the four datasets generated above. The CNN model consists of 2 convolutional layers, 2 pooling layers and a fully-connected layer with a 10-way softmax classifier. The inputs of the CNN are images of size 28×28. The first convolutional layer has 32 kernels with size 5×5, while the other layer has 64 kernels with size 5×5. The Rectified Linear Units (ReLUs) are chosen as the activation function. Every convolutional layer is followed by a non-overlapping max-pooling with filter of size 2×2. In addition, to improve the generalization and convergence performance, a "dropout" layer, which sets the neurons to zero in fully-connected layer with probability of 0.4, is applied before the softmax classifier. Stochastic gradient descent algorithm is employed to optimize the network with a learning rate of 0.001. When the network converges, we evaluate the classification accuracy on the testing set with an average value of 10 times. The last column of Table 1 shows the classification performance comparison of the four generated datasets. As we can see from the Table, if the testing dataset of MNIST is contaminated by the SGWT noise (that is, transformed by SGWT), then the recognition rate can achieve 98.83% when the training dataset of MNIST is augmented by SGWT. However, if the testing dataset of MNIST is contaminated by the SGFRWT, then the recognition rate reduces to 44.55% when the training dataset of MNIST is augmented by SGWT, and increases to 97.44% when using traditional data augmentation methods (flipping, rotating and adding noise), and further increases to 99.01% when using the proposed SGFRWT data augmentation method.

## 6. Conclusion

This paper investigates the issue of extension of spectral graph wavelet transform (SGWT) to fractional domain. The main contributions of this paper can be summarized as follows: (1) A novel transform named spectral graph fractional wavelet transform (SGFRWT) is defined; (2) A Fourier series approximation based fast algorithm for SGFRWT is derived and implemented since the SGFRWT includes complex domain computations compared to SGWT; (3) Applications of SGFRWT to real and synthetic data sets are also given to highlight its potential usefulness.

**Acknowledgement**

This work was supported by the National Key R&D Program of China (2017YFC0107900, 2017YFC0109202), and by the National Natural Science Foundation of China (No. 61201344, 61271312, 61773117, 61401085, 61572258, 11301074, 31400842), and by the Project Sponsored by the Scientific Research Foundation for the Returned Overseas Chinese Scholars, State Education Ministry, by the Qing Lan Project and the '333' project (No. BRA2015288), and by the Short-term Recruitment Program of Foreign Experts (WQ20163200398).

Table 1. Dataset generation implementation details. Comparison of classification performance of four generated datasets on a same CNN network.

| Dataset | Composition Details | Number of dataset images | Accuracy (%) |
| --- | --- | --- | --- |
| (i) | SGWT on MNIST training set, SGWT on MNIST test set. | 360,000 for training, 60,000 for testing. | 98.83 |
| (ii) | SGFRWT on both MNIST training and testing sets. | 1.8 million for training, 0.3 million for testing. | 99.01 |
| (iii) | SGWT on MNIST training set, SGFRWT on MNIST testing set (picking up parts of testing sets). | 360,000 for training, 60,000 for testing. | 44.55 |
| (iv) | SGWT on MNIST training set (and then using traditional data augmentation methods ), SGFRWT on MNIST testing set. | 1.8 million for training, 0.3 million for testing. | 97.44 |

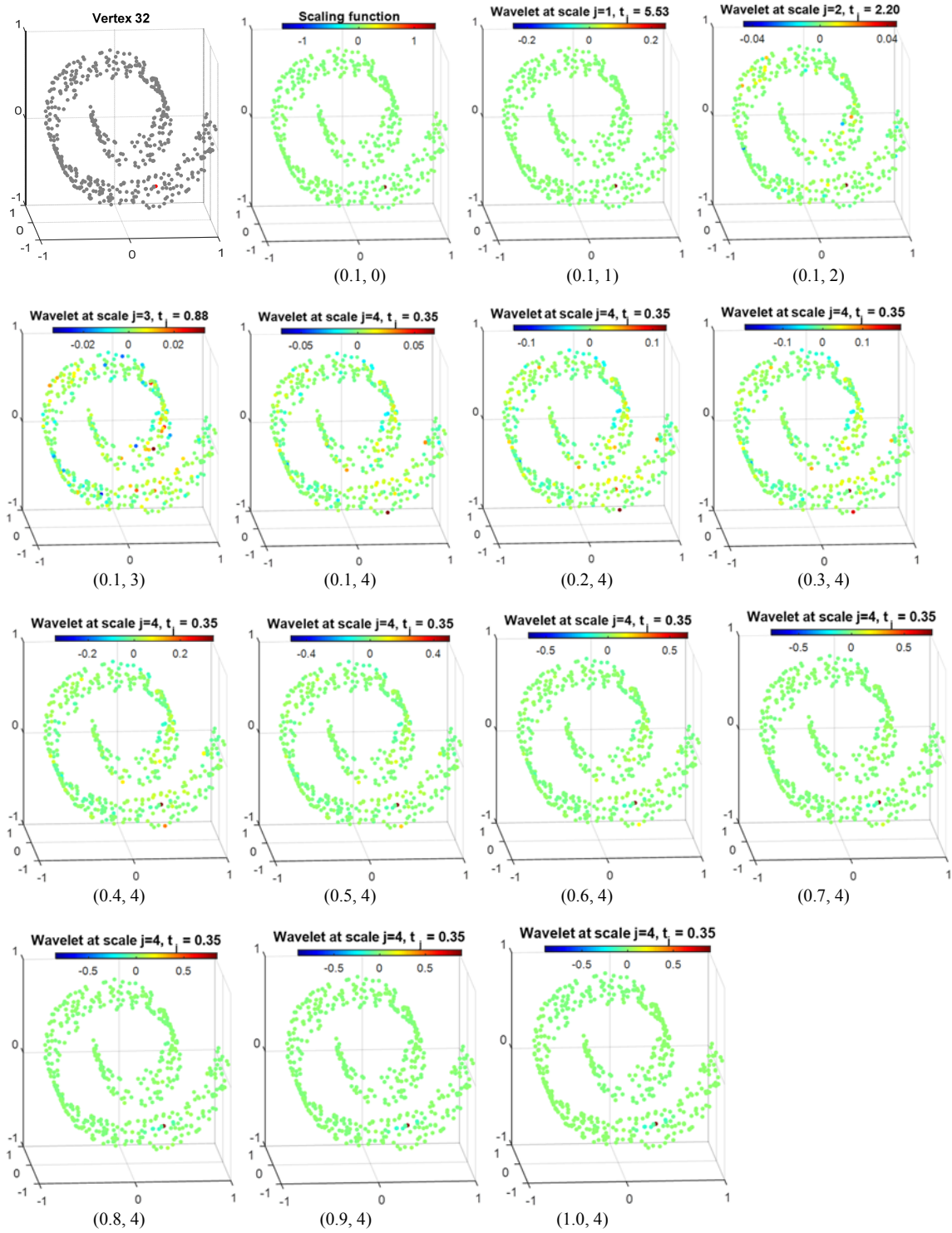

Fig. 1. Spectral fractional graph wavelets on Swiss roll data cloud, with J = 4 wavelet scales and fractional order θ ∈ [0.1,1.0]. The first figure shows vertex at which SGFRWT is centered. (θ, j), θ denotes the fractional order, j denotes the scale (j=0 denotes the scaling function). For example, (0.1, 0) denotes the results of the scaling function of SGFRWT with fractional order 0.1 and (1.0, 4) denotes the results of the j=4 scale of SGFRWT with fractional order 1.0.

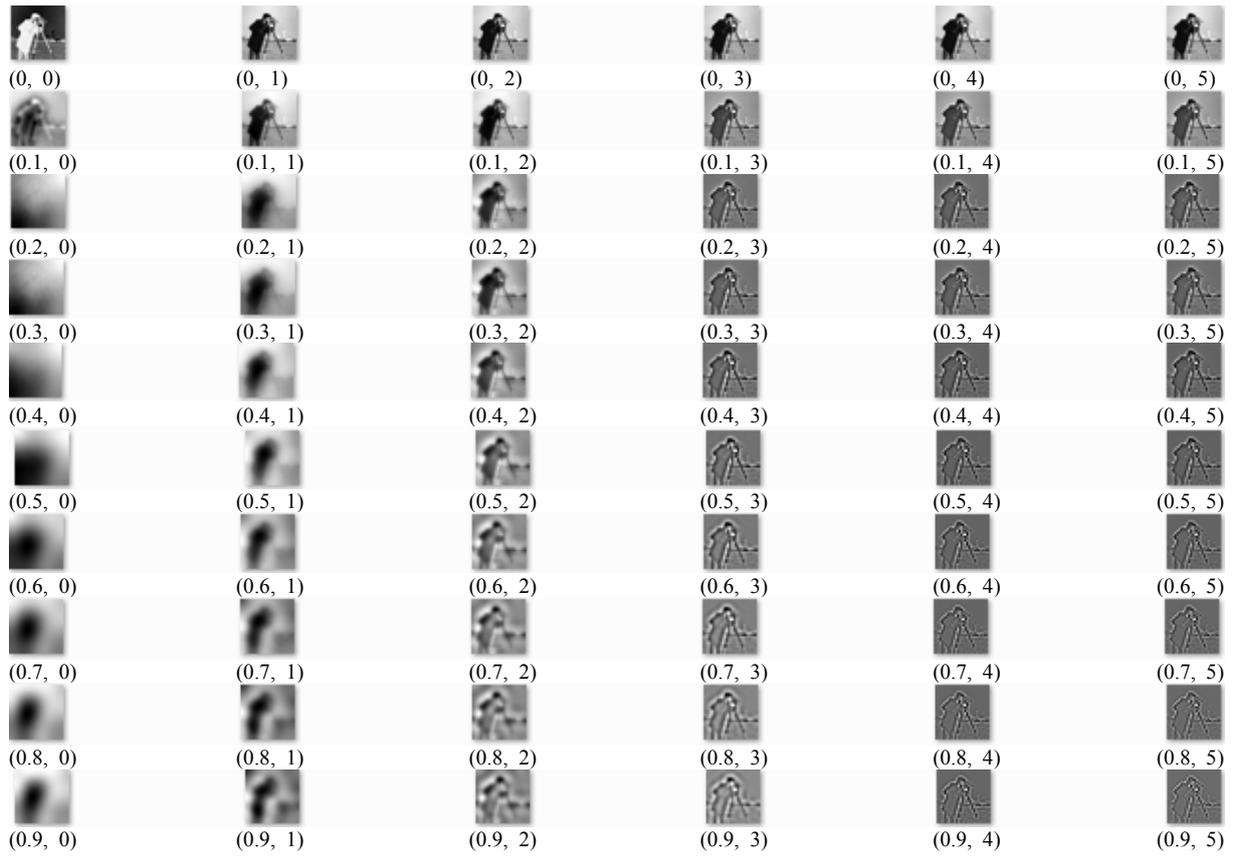

Fig. 2. Spectral graph fractional wavelets on cameraman image, with J = 5 wavelet scales and fractional order θ ∈ [0.0,1.0]. (θ, j), θ denotes the fractional order, j denotes the scale. For example, (0.1, 0) denotes the results of the scaling function of SGFRWT with fractional order 0.1 and (1.0, 5) denotes the results of the *j*=5 scale of SGFRWT with fractional order 1.0.

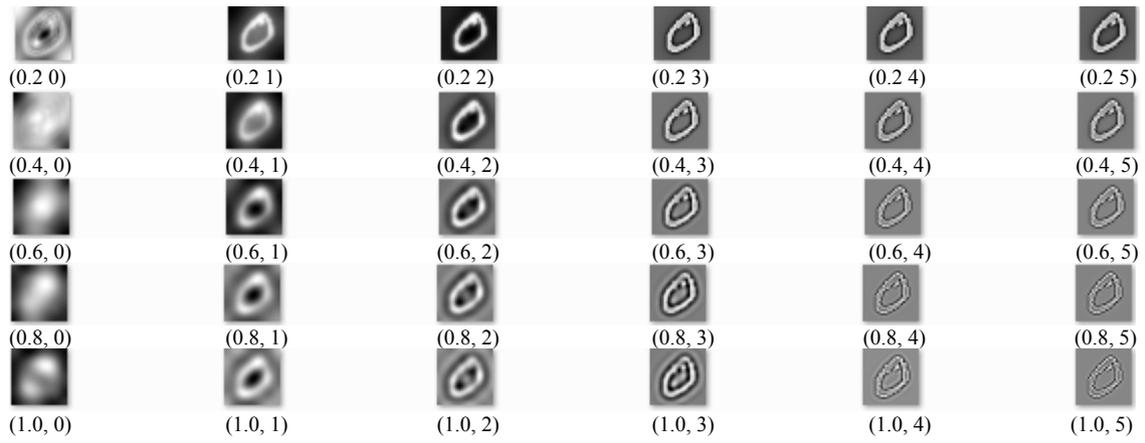

Fig. 3. Spectral graph fractional wavelets on an image "zero", with J = 5 wavelet scales and fractional order θ ∈ {0.2, 0.4, 0.6, 0.8, 1.0}.

(θ, j), θ denotes the fractional order, j denotes the scale.

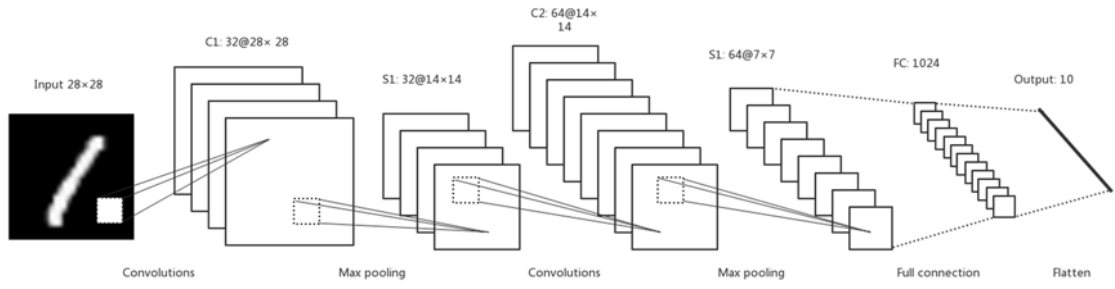

Fig. 4. The convolutional neural network used in our experiment.